# Dhan-Shomadhan: A Dataset of Rice Leaf Disease Classification for Bangladeshi Local Rice

## Authors

Md. Fahad Hossain[1]

## Affiliations

1. Department of Computer Science and Engineering. Daffodil International University

## Corresponding author(s)

Md. Fahad Hossain (fahad15-9600@diu.edu.bd)

## Abstract

This dataset represents almost all the harmful diseases for rice in Bangladesh. This dataset consists of 1106 image of five harmful diseases called Brown Spot, Leaf Scaled, Rice Blast, Rice Turngo, Steath Blight in two different background variation named field background picture and white background picture. Two different background variation helps the dataset to perform more accurately so that the user can use this data for field use as well as white background for decision making. The data is collected from rice field of Dhaka Division. This dataset can use for rice leaf diseases classification, diseases detection using Computer Vision and Pattern Recognition for different rice leaf disease.

## Specifications Table

| | |
|---|---|
| **Subject** | Agriculture |
| **Specific subject area** | Data related to nature, Computer Vision and Pattern Recognition, Leaf diseases classification |
| **Type of data** | Images |
| **How data were acquired** | Capturing photos from local rice field by mobile.<br>Device: Vivo Y15 with Triple 13+8+2 Megapixel camera |
| **Data format** | Raw<br>JPG |

| Parameters for data collection | We tried to parameterize our data by the following disease. Rice blast, Rice Tungro, Sheath Blight, Brown spot and Leaf scald. We focused on the performance of data and take only the data which is clear enough to classify the specific disease so that the data can use as the most accurate one. |
|---|---|
| Description of data collection | We searched for our necessary data in the source location and categorized according to the importance of usefulness and efficiency. We took photos from rice field in different time and weather like foggy morning, light full noon, evening, after rain etc. We concentrate to add every possible important data for maximum usability and performance. |
| Data source location | Dhaka Division<br>Dhaka<br>Bangladesh |
| Data accessibility | Repository name: Mendeley<br>Data identification number: 10.17632/znsxdctwtt.1<br>Direct URL to data: https://data.mendeley.com/datasets/znsxdctwtt/1 |

**Value of the Data**

- This dataset is important in term of rice disease analysis like disease classification or disease detection to track the path of analysing rice product and farmers awareness about these diseases.

- Researchers and scientists can use the dataset to examine the characteristic of these disease and can understand the situation about rice diseases. People especially farmers can see the dataset and increases their knowledge about those disease and take necessary step to stop these diseases.

- This dataset can be used to detect new rice virus. Besides, it can be used to understand the threat of different rice disease. It will be easy to act quickly if any further diseases for rice occurs.

- This dataset may help to keep the track of decreasing rice diseases and its effect. It also can be used to understand the threat of rice diseases in different stages. This dataset also inspires people can take proper action against those diseases.

**Data Description**

In Bangladesh, Rice is the most dominant food crop and most of the arable lands are used for growing rice. So, ensuring the healthy lifespan of paddy is one of the most important tasks for a Bangladeshi farmer. It is also an import liability of researchers to provide necessary information to farmers. So, the data containing this dataset can provide information related to mentioned diseases and can help farmers to identify different disease with corresponding action. Unfortunately, there are no structured dataset about rice leaf disease. Some researchers perform some classification task with a very few numbers of data with white background and it is also possible couting different types of diseases[1-2]. But the most important thing is the presented dataset contain white background image as well as field background image. The benefit is that the user can compare the data in field visit as well as laboratory examination. So, the performance will rise by almost double concerning other researcher's data.

Here are some short description about the disease we classify:

1. **Rice blast:** It is a common rice disease which was caused by a fungus called *magnaporthe grisea*. Blast fungus can infect every part of rice plant including panicles, leaves, neck, nodes, seeds etc. Every growth stages and development period can be infected by it. Entire rice crops failure has resulted from rice blast epidemics. It's a huge threat for rice production [3]. Common symptoms of rice Blast disease are there will be spots in the leaves and light-yellow spots on the side of the leaves. Two sample of Rice blast diseases are given below where fig.1.1 contains field picture and fig.1.2 contains white background picture.

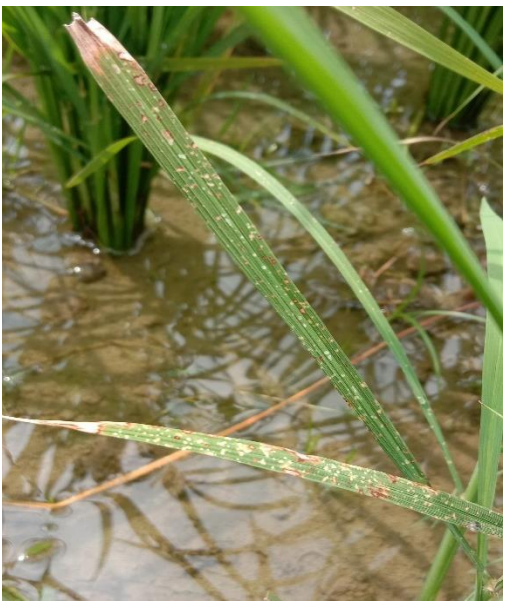
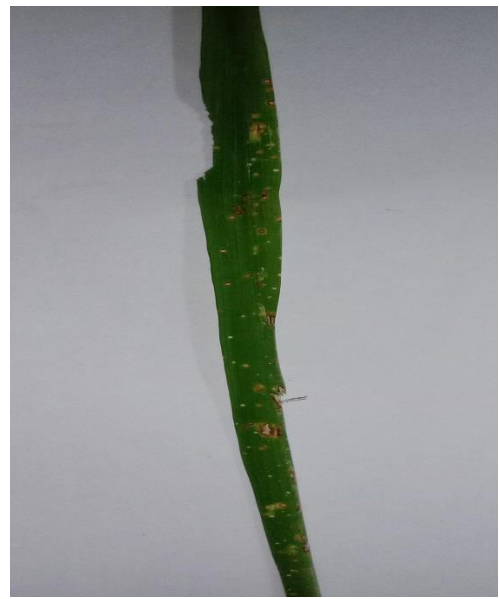

Fig.1.1: Rice Blast in Field             Fig.1.2: Rice Blast in White Background

2. **Rice Tungro:** It is Another common harmful disease for rice. Bacilliform virus (RTBV) and Rice Tungro spherical virus (RTSV) are responsible for tungro diseases. It's not facile to detect this disease from the field because of its different symptoms. Different symptoms including with orange yellow coloring of leaves, stunting, poor panicle emergence, reduced tillering and reduction in tiller number. From this kind of symptoms, we can detect tungro diseases.[4] In every stages of rice growth can be infected by these diseases. Two sample of Rice tungro diseases are given below where fig.2.1 shows field picture and fig.2.2 shows white background picture.

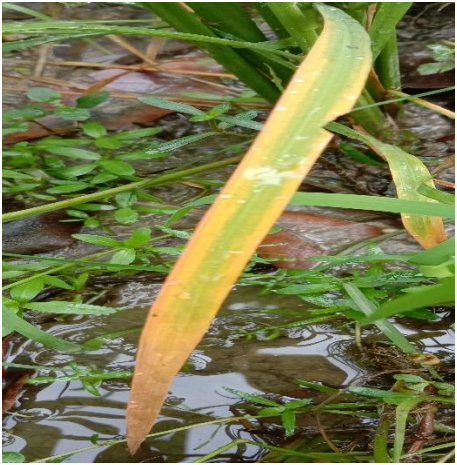
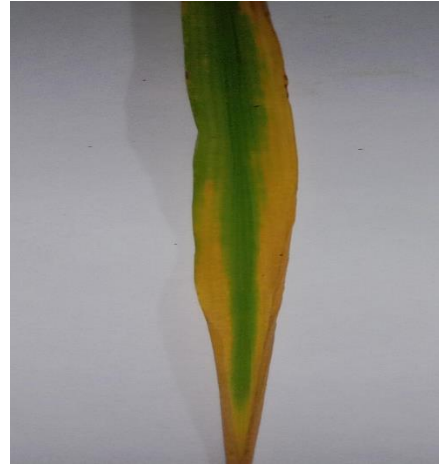

Fig.2.1: Rice tungro in Field                         Fig.2.2: Rice tungro in White Background

3. **Sheath blight:** It is one of the most subversive diseases for rice. It's a fungus affected diseases called *Rhizoctonia Solani.* This fungus affects rice from tillering to heading stage. Primary symptoms are noticed on leaf sheaths near water level. In leaf sheath oval or irregular greenish grey spots are formed. When the spots enlarge, the center becomes greyish white with an irregular blackish brown or purple brown border.[5] Two sample of Sheath blight diseases are given below where fig.3.1 shows field picture and fig.3.2 shows white background picture.

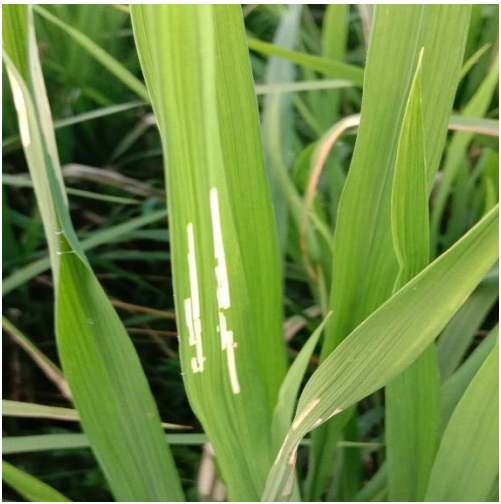
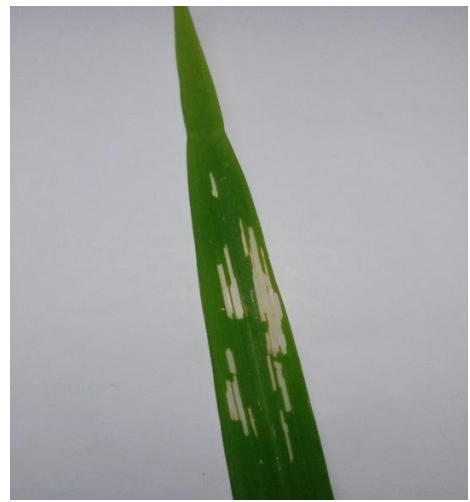

Fig.3.1: Sheath blight in Field                      Fig.3.2: Sheath blight in White Background

4. **Brown spot:** It is another pernicious disease for rice. It's also caused by a fungus. *Cochliobolus miyabeanus* fungus is responsible for brown spot diseases. It is also called Helminthosporium leaf spot, one of the most dominant rice diseases. In its symptoms, some large and small spots are found in rice where smaller spots are dark brown to reddish-brown and larger spots have a dark brown margin and a light reddish-brown or gray center.This fungus also attacks the coleoptiles, branches of the panicle, glumes and grains.[6] Two sample of Brown spot diseases are given below where fig.4.1 shows field picture and fig.4.2 shows white background picture.

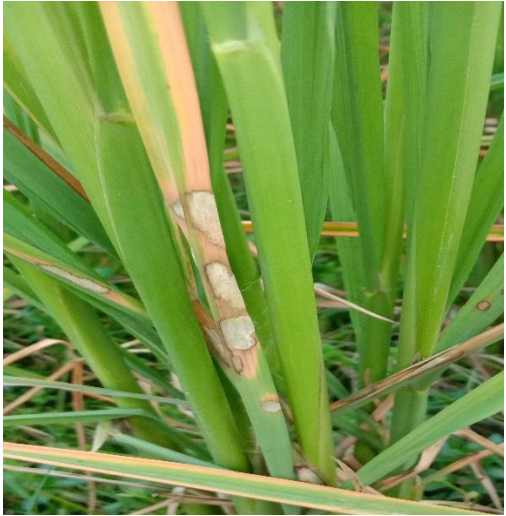 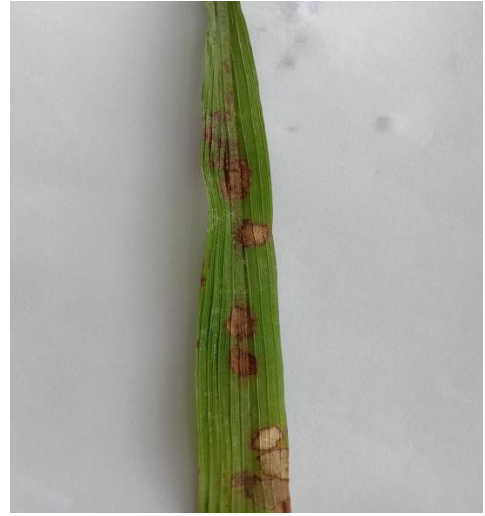

        Fig.4.1: Sheath blight in Field         Fig.4.2: Sheath blight in White Background

5. **Leaf scald**: It is another harmful disease for rice. Microdochium oryzae is responsible for leaf scald. In each case, water-soaked lesions start to build on the edges of leaf. Then lesions increased and produce a ring pattern of light tan and dark brown from the leaf tips.[7] This disease usually occurs on mature leaves of rice which is very harmful for rice production. Two sample of Leaf scald diseases are given below where fig.5.1 shows field picture and fig.5.2 shows white background picture.

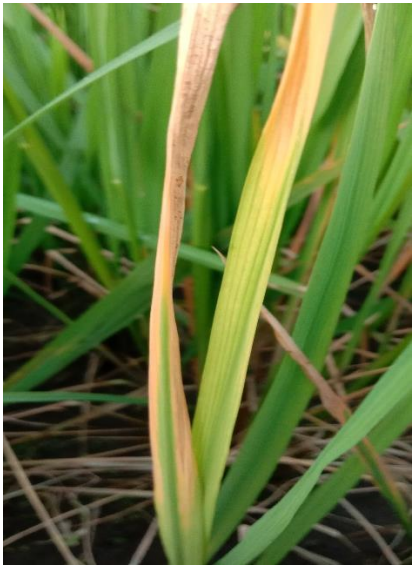 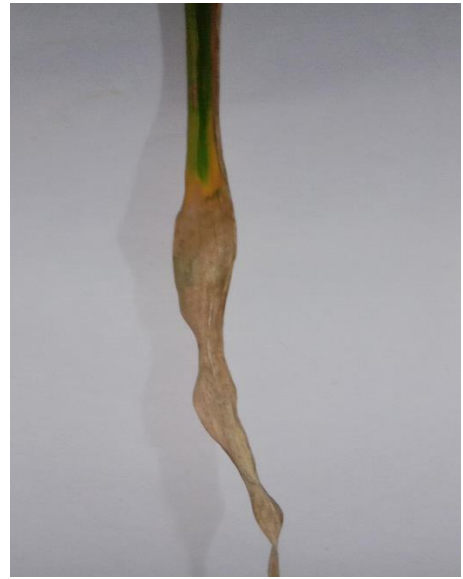

        Fig.5.1: Leaf scald in Field        Fig.5.2: Leaf scald in White Background

Fig.6 shows the number of images for each disease on field background.

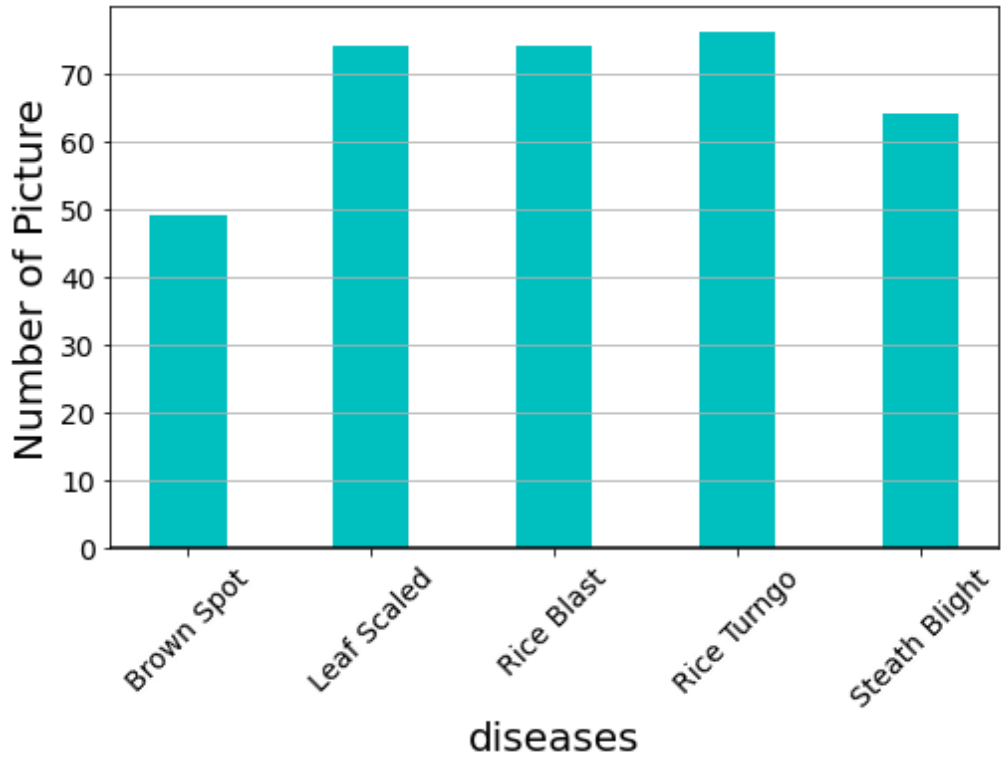

Fig.6: number of images for each disease on field background

Fig.7 shows the number of images for each disease on field background.

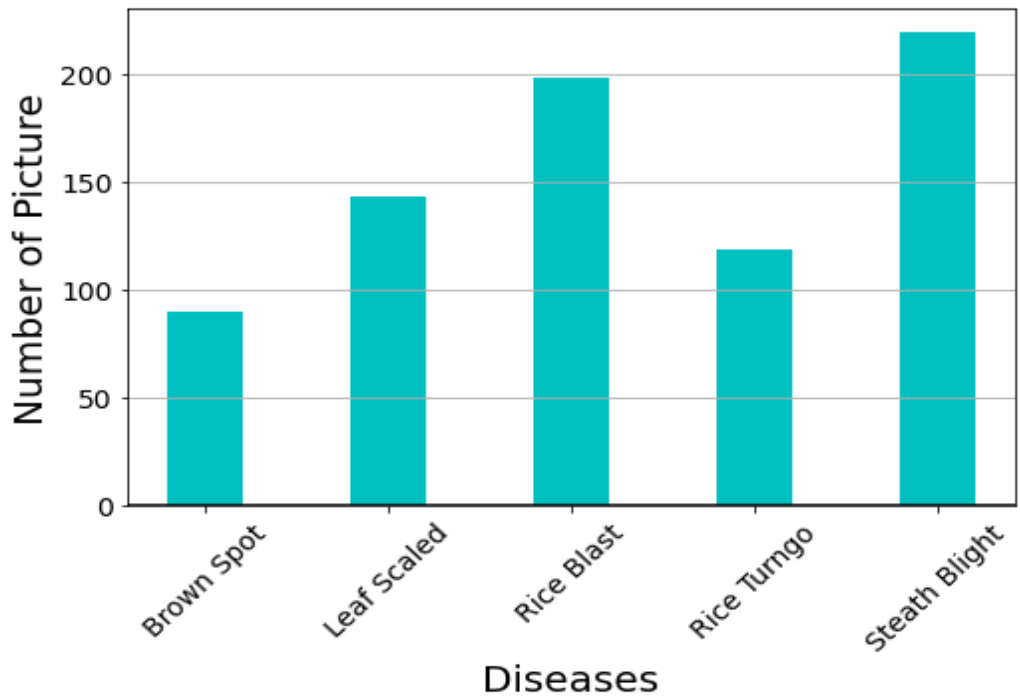

Fig.7: number of images for each disease on white background

**Experimental Design, Materials and Methods**

In Bangladesh, There are three types of rice seasons named Aus, Aman, Boro[8]. Table.1 shows the season duration.

| Season Name | Duration |
|---|---|
| Aus | June-December |
| Aman | March-August |
| Boro | November- March |

In each Seasons there are three phase called vegetative phase, Reproductive phase, Ripening Phase. Table2. Shows the Description of each phase [9].

| Phase Name | Discription |
|---|---|
| Vegetative Phase | The seed grow into a seedling and those seedlings are planted into field and the seedling gradually start growing and ends at tillering which takes Approximately 40-45 days. |
| Reproductive phase | The first sign that the rice plant is getting ready to enter its reproductive phase is a bulging of the leaf stem that conceals the developing panicle. Rice is said to be at the 'heading' stage when the panicle is fully visible. Flowering begins a day after heading has completed. For completing this phase, it needs Approximately 35 days. |
| Ripening Phase | The ripening phase starts at flowering and ends when the rain is mature and ready to be harvested which takes Approximately 35 days. |

Basically, in Reproductive phase these rice diseases can be seen. As Reproductive phase comes after Vegetative phase and before Ripening phase so, we follow Aman and Boro seasons. The Reproductive phase of those seasons are winter and summer in perspective of Bangladesh. So, we need to wait for the reproductive phase to suspect the disease and perform action according to the effect and symptoms. We considered two conditions for capturing the images.

In term of providing most suitable and accurate data, we examine the image concerning different background. Sometimes it is difficult to detect a specific disease in white background or field background. But our dataset contains data with both type of backgrounds. So that it can be detected on both backgrounds. Here, the two types of background we considered while collecting data:

1. **Field Picture:** Image containing field background are collected directly from rice field. These images contain greenish blur background of paddy tree, leaves or fields soil. We consider different time and weather like foggy morning, light full noon, evening, after rain, etc for providing more accurate data. We captured only mentioned five different diseases affected leafs and tried to focus only on the affected spot.

2. **White Background:** Image containing white background are captured in a room with normal sun light. The affected rice leaf are collected from different rice field according to the mentioned five disease class. We considered white paper as the background of white background images.

We used Vivo Y15 with Triple 13+8+2 Megapixel camera. The dimension of each image is 1952*4160. While capturing the images the focal length of the camera was 4mm.

**Declaration of Competing Interest**

The authors declare that they have no known competing financial interests or personal relationships which have or could be perceived to have influenced the work reported in this article.

**References**

[1] M. T. Sami, D. Yan, H. Huang, X. Liang, G. Guo and Z. Jiang, "Drone-Based Tower Survey by Multi-Task Learning," 2021 IEEE International Conference on Big Data (Big Data), Orlando, FL, USA, 2021, pp. 6011-6013, doi: 10.1109/BigData52589.2021.9672078.

[2] M. T. Sami et al., "Center-Based iPSC Colony Counting with Multi-Task Learning," 2022 IEEE International Conference on Data Mining (ICDM), Orlando, FL, USA, 2022, pp. 1173-1178, doi: 10.1109/ICDM54844.2022.00150.

[3] Shahriar, Saleh & Imtiaz, Abdullah & Hossain, Belal & Husna, Asmaul & Nurjahan, Most & Eaty, Khatun. (2020). Review: Rice Blast Disease. Annual Research & Review in Biology. 10.9734/ARRB/2020/v35i130180.

[4] Bunawan, Hamidun & Dusik, Lukas & Bunawan, Siti & Mat Amin, Noriha. (2014). Rice Tungro Disease: From Identification to Disease Control. World Applied Sciences Journal. 31. 1221-1226. 10.5829/idosi.wasj.2014.31.06.610.

[5] Singh, Pooja & Mazumdar, Purabi & Harikrishna, Jennifer & Babu, Subramanian. (2019). Sheath blight of rice: a review and identification of priorities for future research. Planta. 250. 10.1007/s00425-019-03246-8.

[6] Singh, Ram & Sunder, & Agarwal, R.. (2014). Brown spot of rice: an overview. Indian Phytopath. Indian Phytopath.. 201-215.

[7] Tatagiba, Sandro & Rodrigues, Fabrício. (2016). Magnesium decreases the symptoms of leaf scald on rice leaves. Tropical Plant Pathology. 41. 10.1007/s40858-016-0080-x.

[8] http://www.brri.gov.bd/site/page/b9720c2f-2621-429e-be79-75d5c42b7414/-

[9] http://ricepedia.org/rice-as-a-plant/growth-phases.